\documentclass[10pt,twocolumn,letterpaper]{article}

\usepackage{cvpr}              
\usepackage{multirow}
\usepackage{svg}
\usepackage{listings}
\usepackage{colortbl}
\usepackage{xcolor}
\usepackage{float}
\usepackage{bibunits}
\defaultbibliographystyle{ieeenat_fullname}
\defaultbibliography{main} 

\definecolor{cvprblue}{rgb}{0.21,0.49,0.74}
\usepackage[pagebackref,breaklinks,colorlinks,allcolors=cvprblue]{hyperref}
\usepackage[linesnumbered,ruled,vlined]{algorithm2e}

\title{GeoViS: Geospatially Rewarded Visual Search for Remote Sensing \\ Visual Grounding}

\author{
Peirong Zhang$^{1,2,*}$\quad
Yidan Zhang$^{1,2,*}$\quad
Luxiao Xu$^{1,2}$\quad
Jinliang Lin$^{1,2}$\\
Zonghao Guo$^{3,\dagger}$\quad
Fengxiang Wang$^{4}$\quad
Xue Yang$^{5}$\quad
Kaiwen Wei$^{6}$\quad
Lei Wang$^{1,2,\dagger}$\\
$^{1}$Aerospace Information Research Institute, Chinese Academy of Sciences\\
$^{2}$ University of Chinese Academy of Sciences\quad
$^{3}$Tsinghua University \\
$^{4}$National University of Defense Technology\\
$^{5}$Shanghai Jiao Tong University \quad $^{6}$Chongqing University\\
{\tt \small https://github.com/Zhang-Peirong/GeoVis }\\
}

\begin{document}

\maketitle

\begingroup
\renewcommand\thefootnote{}
\footnotetext{* Equal contribution.\quad $\dagger$ Corresponding author.}
\addtocounter{footnote}{-1}
\endgroup

\begin{abstract}
Recent advances in multimodal large language models (MLLMs) have led to remarkable progress in visual grounding, enabling fine-grained cross-modal alignment between textual queries and image regions. However, transferring such capabilities to remote sensing imagery remains challenging, as targets are often extremely small within kilometer-scale scenes, and queries typically involve intricate geospatial relations such as relative positions, spatial hierarchies, or contextual dependencies across distant objects.
To address these challenges, we propose GeoViS, a Geospatially Rewarded Visual Search framework that reformulates remote sensing visual grounding as a progressive search-and-reasoning process. Rather than directly predicting the target location in a single step, GeoViS actively explores the global image through a tree-structured sequence of visual cues, integrating multimodal perception, spatial reasoning, and reward-guided exploration to refine geospatial hypotheses iteratively. This design enables the model to detect subtle small-scale targets while maintaining holistic scene awareness.
Extensive experiments on five remote sensing grounding benchmarks demonstrate that GeoViS achieves precise geospatial understanding and consistently surpasses existing methods across key visual grounding metrics, highlighting its strong cross-domain generalization and interpretability.
\end{abstract}

\section{Introduction}
\begin{figure}[t!]
  \centering
  \includegraphics[width=\linewidth]{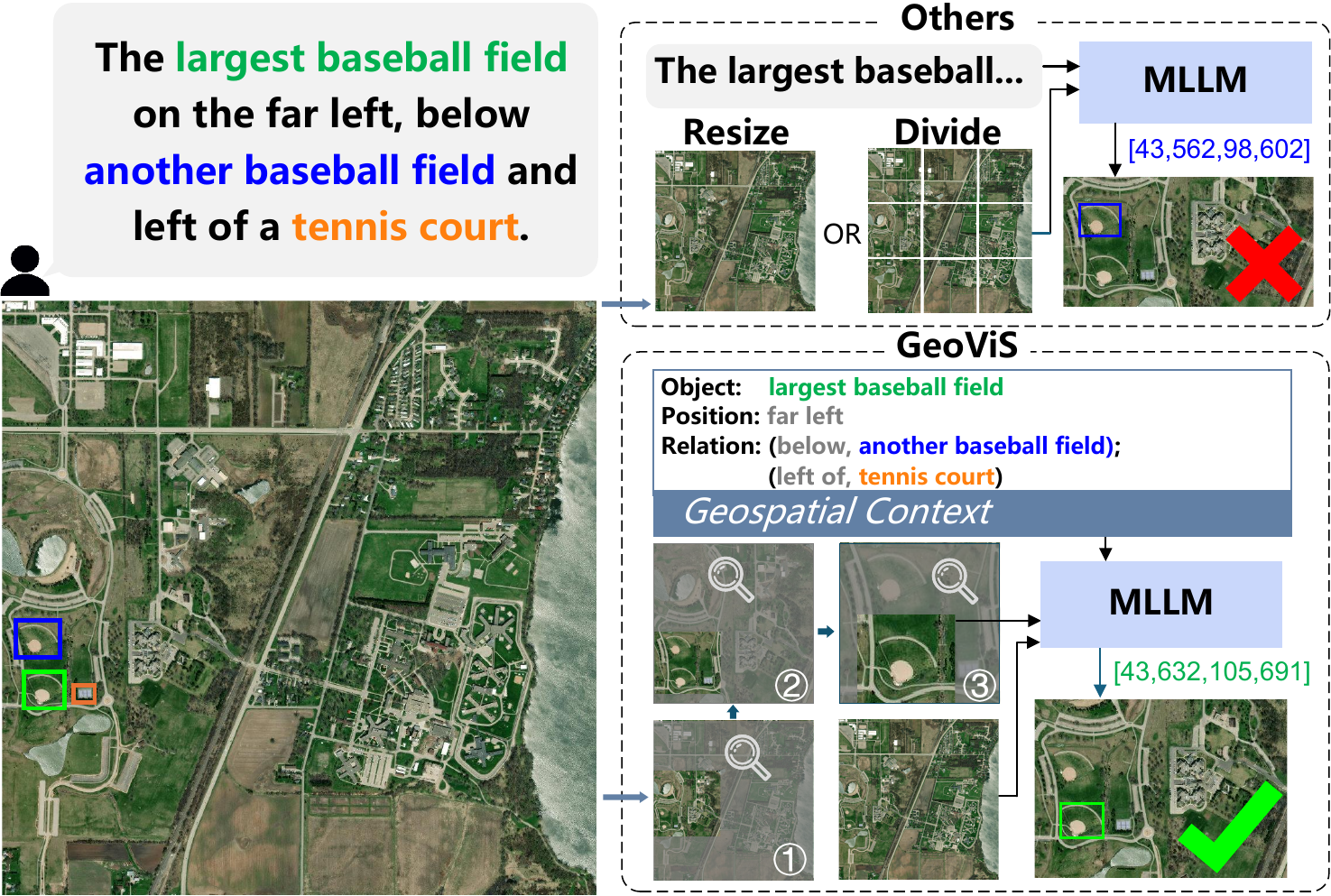}
  \caption{Complex queries with multi-object relations and tiny targets make remote sensing grounding challenging. While existing one-step methods that resize or divide images often fail, GeoViS parses structured semantics and performs reward-guided subregion exploration to achieve accurate localization.}
  \label{fig:Intro}
\end{figure}

Recent advances in multimodal large language models (MLLMs)\cite{guo2024llava,Qwen2.5-VL,liu2024improved,alayrac2022flamingo}
have significantly improved cross-modal alignment and reasoning.
Visual grounding\cite{deng2021transvg,yang2019fast,yang2020improving,sadhu2019zero,li2021referring}
, which aims to localize image regions based on textual queries, is a fundamental ability for vision–language understanding\cite{you2023ferret,chen2023shikra,li2022grounded,yang2023dawn,chen2024internvl,wang2024cogvlm}. 
Nevertheless, extending this capability to remote sensing imagery remains highly challenging due to its unique characteristics\cite{kuckreja2024geochat}.

First, the top-down perspective of remote sensing imagery introduces complex spatial relationships. 
Textual queries in remote sensing often involve multiple entities and complex spatial relations, requiring the model to reason over object attributes and their relative positions rather than recognizing isolated instances. 
As illustrated in Fig.~\ref{fig:Intro}, for example, the model must first identify all baseball fields, compare their relative sizes to determine the largest one, and then reason about positional relations with surrounding objects such as the tennis court to accurately localize the final target.
However, existing approaches~\cite{peng2023kosmos,li2024groundinggpt,ma2024groma} typically perform one-shot prediction over the entire image, lacking hierarchical spatial modeling and thus struggling with such complex queries, often leading to inaccurate localization.
Although multi-step reasoning techniques based on chain-of-thought or reinforcement learning have been explored in natural domains~\cite{li2025dyfo,shao2024visual,lai2025mini,zheng2025deepeyes,su2025pixel,zhang2025chain}, they usually demand substantial human involvement to construct large-scale reward datasets for training. Such approaches are therefore complex and costly, limiting their applicability in remote sensing scenarios. 

Second, remote sensing images typically span kilometer-scale areas, while objects of interest such as aircraft or ships occupy only a small fraction of the scene~\cite{hong2019patch,wu2022fsanet}.
Compared with natural images that are object-centric and resolution-balanced, remote sensing scenes suffer from extreme scale disparity between the target and the background. 
This results in much lower \textit{effective resolution}, defined as the proportion of pixels representing the target relative to the whole image. 
Existing methods\cite{zhou2024geoground,soni2025earthdial,muhtar2024lhrs} mainly attempt to increase image resolution by enlarging the input or dividing the image into patches, 
yet these strategies do not change the effective resolution\cite{shi2025scaling}.
Consequently, small objects lose discriminative visual details, making accurate localization particularly difficult.

Our preliminary experiments (see Sec.~\ref{sec:preExp}) show that grounding performance improves substantially when the model can (i) correctly interpret the geospatial context expressed in the textual description, and (ii) incorporate a candidate region that is likely to contain the target as an additional visual input to increase the effective resolution.
We refer to these two components jointly as visual cues.
Automatically discovering reliable visual cues in large and cluttered remote sensing scenes, however, remains the key challenge for accurate grounding.

To address these challenges and enhance grounding capability in remote sensing MLLMs, we introduce GeoViS, a \textbf{Geo}spatially rewarded \textbf{Vi}sual \textbf{S}earch framework that reformulates visual grounding as a multi-step reasoning and exploration process.
GeoViS navigates large-scale remote sensing imagery using a geospatial reward that quantifies how well each region satisfies the visual cues derived from both geospatial context and progressively refined regional evidence, thereby guiding the search toward promising locations.
Specifically, it decomposes spatial queries into interpretable sub-problems and performs hierarchical reasoning through a reward-driven progressive search procedure over candidate regions.
To support this process, we develop a unified Visual Reward–Action–Grounding (VisualRAG) model, which evaluates geospatial rewards, guides region-level actions through multimodal feedback, and performs precise grounding once the target region is identified.
Through the joint modeling of geospatially rewarded search and multimodal grounding, GeoViS uncovers reliable visual evidence and achieves accurate localization for targets with complex geospatial context.

Extensive experiments across five remote sensing grounding benchmarks~\cite{zhan2023rsvg,li2024vrsbench,kuckreja2024geochat,li2024optrsvg,2024Lanrsvghr} demonstrate that GeoViS consistently surpasses both general-purpose and remote sensing specific MLLMs, yielding 5–15\% improvements on key localization metrics and establishing new State-Of-The-Art (SOTA) performance on all datasets. 
The unified training scheme further enables strong cross-dataset generalization, ensuring robust accuracy under diverse scenes and query types.

Our main contributions are summarized as follows:
\begin{itemize}
    \item We propose \textbf{GeoViS}, a geospatially rewarded visual search framework that performs hierarchical tree-structured exploration and conditional grounding driven by geospatial rewards derived from visual cues.
    
    \item We design a unified Visual Reward–Action–Grounding (\textbf{VisualRAG}) model that infers geospatial rewards, guides region-level search actions, and enables fine-grained conditional grounding.
    
    \item We validate GeoViS on five remote sensing grounding benchmarks, showing SOTA performance and robust cross-dataset generalization of the visual search process.
\end{itemize}

\section {Related Work}
\subsection{Remote Sensing Visual Grounding}
Remote sensing visual grounding has drawn increasing attention due to the need for fine-grained cross-modal localization in large-scale scenes. Early methods typically adopt a two-tower paradigm with separate vision and text encoders coupled by cross-modal interaction modules \cite{choudhury2024crossvg,li2024injecting,ding2025visual,ding2024visual,wang2024multistage,zhao2024spatial}. Subsequent works further improve grounding performance through enhanced architectural designs, more expressive interaction mechanisms, or by rethinking the task pipeline and decoding strategies for better efficiency and scalability \cite{zhao2025context,hang2024regionally}.

With the rapid progress of multimodal large language models (MLLMs) \cite{hurst2024gpt,comanici2025gemini,liu2024improved,alayrac2022flamingo}, several RS-specific MLLMs have been introduced \cite{zhang2024earthgpt,wang2024ringmogpt,luo2024skysensegpt,ou2025geopix,pang2025vhm}, showing promising results on RS perception and grounding tasks. However, the extremely low effective resolution of targets within kilometer-scale images remains a fundamental bottleneck: tiny objects become heavily blurred after global encoding, causing substantial detail loss and frequent localization failures in MLLMs. This intrinsic scale disparity highlights the need for models that can progressively reason over large areas while preserving local discriminative cues.

\subsection{Visual Search for MLLMs}
To enhance the fine-grained perception of MLLMs, recent works introduce visual search mechanisms~\cite{tong2024eyes,wu2025f,shao2024visual,li2024vocot,wu2024v,sun2024visual}, enabling models to inspect localized regions instead of relying on a single global encoding. V*~\cite{wu2024v} treats search as a sequence of visuomotor operations (e.g., crop, zoom, shift) predicted by the LLM, while DyFo~\cite{li2025dyfo} proposes a training-free MCTS-based focusing strategy guided by an external detector. However, both methods depend on strong detectors, which are difficult to deploy or generalize in remote sensing, causing error accumulation when targets are small or ambiguous.

Another direction constructs visual chain-of-thought (CoT) data to train models to reason about spatial details~\cite{li2024vocot,tong2024eyes,sun2024visual}. GPT-o3 further demonstrates that dynamic operations such as adaptive scaling greatly benefit small-object recognition. Following this idea, DeepEye~\cite{zheng2025deepeyes} learns an RL policy to sequentially select informative glimpses, and Mini-o3~\cite{lai2025mini} distills o3-like zooming behaviors. Yet these RL-based approaches require large curated training sets and costly rollouts, making them hard to scale or adapt to remote sensing, where high-quality data are limited.

In this work, we adopt a visual search perspective without relying on RL or external detectors. We express the search as a sequence of simple atomic operations with a reward-driven evaluation scheme, enabling efficient data construction from existing remote sensing benchmarks and endowing the model with robust fine-grained localization ability.

\section{Methodology}
\begin{figure*}[h]
  \centering
  \includegraphics[width=\linewidth]{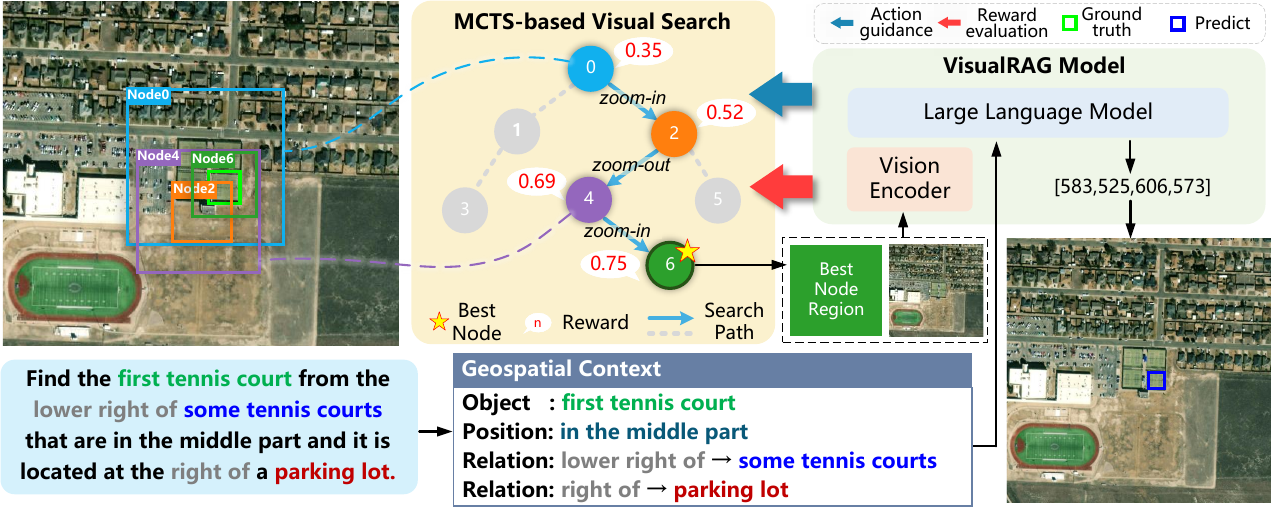}
  \caption{Overview of GeoViS. Complex queries are structured into object, position, and relation cues. GeoViS first performs MCTS-based visual search to identify the most informative subregion, where each node represents a candidate region, and then conducts conditional grounding using the global image and the selected subregion. The VisualRAG model supports the entire pipeline by providing action guidance, reward evaluation, and final localization.}
  \label{fig:methods}
\end{figure*}
  
\subsection{Overview}
Given a remote sensing image $I$ and a textual description $T$, the goal of visual grounding is to localize the target region by predicting its bounding box $B$. 
Such a formulation ignores the progressive reasoning process humans naturally adopt when searching for small targets within large-scale imagery, leading to suboptimal localization under extreme scale and context diversity.
Conventional methods typically treat this as a one-step regression or attention prediction task, which is insufficient for complex remote sensing scenes containing vast search spaces and many visually similar objects. 

We reformulate the task as a geospatially rewarded visual search problem and propose \textbf{GeoViS}, which decomposes grounding into two sequential stages: \emph{visual search} and \emph{visual grounding}.
The search process is guided by geospatial rewards that jointly evaluate semantic alignment and spatial consistency, encouraging the model to explore regions with higher textual relevance and geographic plausibility.
As shown in Fig.~\ref{fig:methods}, in the first stage, GeoViS performs a reward-guided hierarchical search over the global image to progressively locate the subregion that most likely contains the object described by the text.
In the second stage, it conducts conditional grounding within the global image, using the searched subregion as a visual cue to refine the final localization.

To support both stages in a unified manner, we introduce the Visual Reward–Action–Grounding (\textbf{VisualRAG}) model as the core perception and reasoning module of GeoViS.
The VisualRAG model is a multimodal large language model that jointly provides: 
(1) reward evaluation to assess the semantic and geometric consistency of candidate regions, (2) action guidance to predict the optimal zoom-in action for MCTS-based visual search, and (3) grounding reasoning to estimate the target position in the global image. 
By sharing this model across search and grounding, GeoViS achieves coherent, step-wise reasoning over geospatial imagery and effectively bridges exploration, verification, and final localization.

\subsection{Visual Search}
\subsubsection{Problem Formulation}
The visual search process is inherently sequential and decision-driven, where the model must iteratively explore potential regions and refine its spatial hypotheses based on textual cues.
We formulate the visual search process in GeoViS as a Markov Decision Process (MDP) 
$\mathcal{M} = (\mathcal{S}, \mathcal{A}, \mathcal{T}, \mathcal{R})$, 
where $\mathcal{S}$ is the state space, $\mathcal{A}$ the action space, 
$\mathcal{T}$ the state transition function, and $\mathcal{R}$ the reward function. 
At each step $t$, the state $s_t \in \mathcal{S}$ represents a candidate region within the global image, 
characterized by its spatial coordinates and multimodal features. 
An action $a_t \in \mathcal{A}$ transforms the region to a new state according to
\begin{equation}
s_{t+1} = \mathcal{T}(s_t, a_t),
\end{equation}
and the grounding model evaluates this region to produce a scalar reward
\begin{equation}
r_t = \mathcal{R}(s_t, a_t),
\end{equation}
reflecting its semantic relevance to the textual description and guiding the search policy toward more text-consistent regions.

This formulation allows GeoViS to leverage Monte Carlo Tree Search (MCTS) for reasoning-based exploration, balancing exploration of new regions and exploitation of promising hypotheses.

During \textit{selection}, the child node is chosen according to the UCT rule:
\begin{equation}
a^* = \arg\max_{a \in \mathcal{A}(s)} \Big[ Q(s,a) + c \sqrt{\frac{\ln N(s)}{N(s,a) + \varepsilon}} \Big],
\end{equation}
where $Q(s,a)$ denotes the average return of action $a$ at state $s$, 
$N(s)$ and $N(s,a)$ are the visit counts of the node and edge, respectively, 
and $c$ balances exploration and exploitation. 
Next, \textit{expansion} applies an action $a_t \in \mathcal{A}$ to generate a new node, 
followed by \textit{simulation}, where the grounding model evaluates the region to produce a reward $r_t$. 
Finally, \textit{backpropagation} propagates the cumulative reward upward to update the values of all visited nodes. 


Through repeated rollouts, the search tree gradually converges toward regions with higher expected rewards, effectively narrowing the search space from global to local in a self-guided manner.
The following sections describe the design of the action space and reward function in detail.

\subsubsection{Action Space}
Unlike natural image grounding, remote sensing queries often contain compositional semantics—describing objects through attributes and spatial relations rather than explicit names.
To provide explicit semantic cues for spatial reasoning, the textual query $T$ is first transformed into a structured representation that encodes its geospatial context
\begin{equation}
\hat{T} = \Phi(T) = \{ o, p, r \},
\end{equation}
where $o$ denotes the target object, $p$ its spatial attributes, and $r$ the relational references to other entities. 
This structured representation provides interpretable guidance for region-level actions, enabling the model to associate each component of the geospatial context with corresponding visual transitions during search.

The action space in GeoViS is designed to mimic a human’s coarse-to-fine search behavior and is defined as
$\mathcal{A} = \{ \mathcal{T}_{\text{in}}(s_t, a_t), \mathcal{T}_{\text{out}}(s_t, a_t) \}$,
consisting of two complementary operations: \textit{zoom-in} and \textit{zoom-out}.
At each state $s_t$, \textit{zoom-in} performs local refinement by partitioning the current region into a $3 \times 3$ grid and selecting one subregion guided by $\hat{T}$. 
This operation allows the model to focus on finer subregions that potentially contain the target, guided by the textual semantics in $\hat{T}$. 
Formally, if the current region is defined by coordinates $(x_1, y_1, x_2, y_2)$, the new state is obtained as
\begin{equation}
s_{t+1} = \mathcal{T}_{\text{in}}(s_t, a_t) = R_{i,j}(s_t),
\end{equation}
where $R_{i,j}$ denotes the selected subregion among the grid cells.

Conversely, \textit{zoom-out} enlarges the current region by a fixed scaling factor $\lambda > 1$, yielding a broader search scope:
\begin{equation}
s_{t+1} = \mathcal{T}_{\text{out}}(s_t, a_t) = \lambda \cdot s_t.
\end{equation}
This step helps the model recover contextual information when previous exploration becomes overly localized, maintaining awareness of the global scene structure.

Both actions are evaluated at each node to generate new candidate states, allowing the search tree to hierarchically balance local exploration and global context reasoning.

\subsubsection{Reward Design}

In GeoViS, the reward function $\mathcal{R}(s_t, a_t)$ serves as the evaluation signal for each simulated node, corresponding to the \textit{simulation} step in MCTS. 
It measures how well the region represented by state $s_t$ aligns with the textual description. 
The overall reward is composed of two components.

The first term is a \textit{Question–Answering reward}, which evaluates semantic consistency between the candidate region and the linguistic description at a conceptual level.
Given the global image $I_g$, the candidate region $I(s_t)$, and the geospatial context $\hat{T}$, the VisualRAG model verifies each semantic element, including the target object $o$, spatial attribute $p$, and relational reference $r$, through binary predictions.
The QA reward is defined as the normalized proportion of positive responses, denoted by $r_{\mathrm{QA}} \in [0,1]$.

The second term is an \textit{IoU-based reward}, complements the QA term by providing geometric supervision for spatial alignment.
Intuitively, it encourages the search to center on regions where the predicted object occupies a plausible and compact position relative to the region boundary. 
The VisualRAG model is further used to predict the bounding box $B_t$ of the main object $o$ within $I(s_t)$, 
and the predicted box is compared with a virtual central region $B_c$ 
(centered in $I(s_t)$ with half of its width and height) to compute the intersection-over-union score $r_{\mathrm{IoU}}$.

The final reward combines these two terms through weighted normalization:
\begin{equation}
r_t = \alpha\, r_{\mathrm{QA}} + (1 - \alpha)\, r_{\mathrm{IoU}}, \quad r_t \in [0,1],
\end{equation}
where $\alpha$ controls the relative contribution of semantic verification and geometric consistency.
This joint design ensures that both semantic correctness and spatial precision contribute to the search guidance, forming a balanced geospatial reward that drives effective exploration.

\subsection{Visual Grounding}
After the visual search process returns the node $s^\star$ with the highest cumulative reward, the goal of visual grounding is to precisely localize the target region in the global image $I_g$ given the textual description $T$. 
Different from conventional grounding that directly predicts from $(I_g, T)$, 
GeoViS performs a \emph{conditional grounding} process guided by the candidate region $I(s^\star)$ obtained from search:
\begin{equation}
B = \mathcal{G}\!\left(I_g,\, T \mid I(s^\star)\right),
\label{eq:grounding_cond}
\end{equation}
where the subregion $I(s^\star)$ serves as a prior regional visual cue that constrains the search space and enhances the effective resolution of the target area. 
By conditioning on this candidate region, the model can focus on visually and semantically relevant details, allowing it to better perceive the target and accurately project its location back to the global image.

\subsection{VisualRAG Model}
\label{sec:VCR}
The Visual Reward–Action–Grounding (VisualRAG) Model serves as the core perception and reasoning component of GeoViS, performing multimodal understanding over image regions and textual cues. 
Given a global image $I_g$, a candidate region $I(s_t)$, and the structured textual description $\hat{T}=\{o,p,r\}$, the model follows a standard multimodal large language model (MLLM) architecture consisting of an image encoder, a projection layer, and a large language model backbone.

The VisualRAG model unifies three fundamental capabilities that support both the visual search and grounding processes: 
\textbf{(1) Reward Evaluation} evaluates both the semantic correctness of the object–relation triplet $(o,p,r)$ and the spatial consistency of the detected region, producing a unified reward signal that quantitatively reflects how well a candidate region satisfies the textual description; and 
\textbf{(2) Action Guidance} evaluates candidate subregions and infers the optimal action $a_t\in\{1,\ldots,9\}$ for the next-step exploration, serving as the policy signal in the MCTS-based search.
\textbf{(3) Grounding Reasoning} predicts the target position within the global image $I_g$ based on the current region $I(s_t)$ and the textual description $\hat{T}$, enabling accurate localization under multimodal guidance; 

During training, the VisualRAG model is supervised through four atomic operations:
(1) QA reward: semantic verification through question answering;
(2) IoU reward: spatial supervision from object-level localization;
(3) Zoom-in action: hierarchical region selection via $3\times3$ grid partitions; and
(4) Conditional grounding: predicting the target box using both the global image and the selected region as joint inputs.
These components jointly enable the model to integrate reward estimation, action prediction, and grounding within a unified framework.

During inference, the model interacts with the MCTS-based visual search to guide region exploration and scoring.  
Once the search converges to the best node $s^\star$, the corresponding subregion $I(s^\star)$ and the global image $I_g$ are jointly fed into the VisualRAG model for the final conditional grounding, yielding the predicted bounding box $B^\ast$.

\section{Experiments}

\begin{table*}[t]
\centering
\setlength{\tabcolsep}{8pt}
\begin{tabular}{l|ccc|ccc|c} 
\toprule
\multirow{2}{*}{\textbf{Methods}} &
\multicolumn{3}{c|}{\textbf{DIOR-RSVG}} &
\multicolumn{3}{c|}{\textbf{VRSBench}} &
\multicolumn{1}{c}{\textbf{GeoChat}} \\
\cmidrule(lr){2-4} \cmidrule(lr){5-7} \cmidrule(lr){8-8}
& \textbf{Pr@0.5} & \textbf{Pr@0.7} & \textbf{meanIoU} &
\textbf{Pr@0.5} & \textbf{Pr@0.7} & \textbf{meanIoU} &
\textbf{Pr@0.5} \\
\midrule
\multicolumn{8}{l}{\textbf{Specialist Models}} \\
\midrule
\color{gray}
LQVG\cite{2024Lanrsvghr}& \textcolor{gray}{83.4} & \textcolor{gray}{75.9} & \textcolor{gray}{74.0} & - & - & - & -  \\

\color{gray}
LPVA\cite{li2024optrsvg} & \textcolor{gray}{82.8} & \textcolor{gray}{72.3} & \textcolor{gray}{72.4} & - & - & - & -  \\
\midrule
\multicolumn{8}{l}{\textbf{General VLMs / General MLLMs}} \\
\midrule

Claude-sonnet-4 * & 17.6 & 1.2 & 25.3 & 11.1 & 2.4 & 16.7 & -  \\
Gemini-2.0-flash * & 20.8 & 3.3 & 27.5 & 22.9 & 6.3 & 28.6 & -  \\
ChatGPT-5 * & 26.1 & 3.3 & 28.4 & 14.4 & 2.3 & 22.7  & -  \\
MiniGPT-v2\cite{chen2023minigpt} & 29.4 & 10.2 & 29.4 & 32.1 & 16.3 & 34.0 & -  \\
Qwen2.5-VL-3B\cite{Qwen2.5-VL} & 33.4 & 20.0 & 33.8 & 37.3 & 20.8 & 35.3 & 10.3  \\
GLM-4.1V-9B-Thinking\cite{hong2025glm} & 47.6 & 33.8 & 45.3 & 43.4 & 35.7 & 48.1 & 16.4  \\
Qwen3-VL-4B\cite{qwen3vl2025} & 51.1 & 38.1 & 47.8 & 56.3 & \underline{36.1} & 49.8 & 17.1  \\
\midrule
\multicolumn{8}{l}{\textbf{RS MLLMs}} \\
\midrule
EarthDial\cite{soni2025earthdial} & 46.1 & 30.2 & 39.5 & 14.4 & 7.8 & 13.0 & -  \\
VHM\cite{pang2025vhm} & 55.9 & 35.5 & 49.9 & 33.9 & 10.0 & 34.9 & -  \\
Geochat\cite{kuckreja2024geochat} & 56.3 & 24.6 & 53.5 & 31.4 & 11.0 & 35.0 & \underline{22.7}  \\
SkySenseGPT\cite{luo2024skysensegpt} & 60.8 & 26.5 & 53.2 & \underline{63.5} & 26.0 &\underline{54.6} & - \\
EarthGPT\cite{zhang2024earthgpt} & 76.7 & \underline{66.5} & \underline{72.4} & - & - & - & -  \\
GeoGround\cite{zhou2024geoground} & \underline{77.7} & - & - & 49.8 & 20.0 & - & -  \\
\midrule
\textbf{GeoViS(Our)} & \textbf{79.8} & \textbf{70.1} & \textbf{72.6} & \textbf{68.5} & \textbf{45.7} & \textbf{59.2} & \textbf{23.7}  \\
\bottomrule
\end{tabular}
\caption{Results on DIOR-RSVG, VRSBench, and GeoChat. GeoViS is trained on Qwen2.5-VL-3B. Best and second-best results are bolded and \underline{underlined}. Blank entries denote unreported metrics, and models marked with * use evaluation results sourced from \cite{liu2025faithfulreasoningremotesensing}.}
\label{tab:dior_vrs_geo_results}
\end{table*}

\subsection{Datasets}

\textbf{DIOR-RSVG}~\cite{zhan2023rsvg} serves as a foundational benchmark for identifying salient objects within remote sensing scenes. It comprises 38,320 samples characterized by a fixed image resolution of $800\times800$ pixels, providing a standardized setting for evaluating baseline grounding capabilities.

 \textbf{RSVG-HR}~\cite{2024Lanrsvghr} is utilized to assess the model's capability in locating inconspicuous targets within high-resolution imagery. This dataset offers a focused collection of 2,650 samples, all maintained at a consistent resolution of $1024\times1024$ pixels to challenge fine-grained object detection.

 \textbf{OPT-RSVG}~\cite{li2024optrsvg} evaluates model robustness against variable observation conditions and image scales. By integrating three distinct data sources into a unified set of 48,952 samples with varying image dimensions, it challenges models to adapt to diverse spatial resolutions.

 \textbf{VRSBench}~\cite{li2024vrsbench} provides a visual grounding subset designed for fine-grained reasoning tasks. Utilizing standardized $512\times512$ pixel patches with high-quality, human-verified annotations, it is partitioned into 36,307 training and 16,159 testing samples to ensure rigorous evaluation.

 \textbf{GeoChat}~\cite{kuckreja2024geochat} provides a remote-sensing multimodal dataset spanning resolutions from $600\times600$ to larger scales. We filter it to isolate single-object grounding tasks, yielding 66,330 training and 1,890 testing samples for evaluating performance on noisy, auto-generated instructions across heterogeneous image sizes.

\subsection{Metrics} 

We evaluate the performance of visual grounding using Precision at IoU thresholds (Pr@0.5 and Pr@0.7) and the mean Intersection over Union (meanIoU). The meanIoU quantifies the average overlap between the predicted and ground-truth bounding boxes, and is defined as: 
\begin{equation} \text{meanIoU} = \frac{1}{M} \sum_{t=1}^{M} \frac{I_t}{U_t}, \end{equation} 
where $M$ denotes the total number of image–query pairs, and $I_t$ and $U_t$ represent the intersection and union areas of the predicted and ground-truth boxes, respectively.

\subsection{Implementation Details}
The training is conducted on 8 NVIDIA A800 GPUs using the LLaMA-Factory framework \cite{zheng2024llamafactory}.
The VisualRAG model is initialized from Qwen2.5-VL-3B-Instruct~\cite{Qwen2.5-VL} and fine-tuned in a fully-trainable manner, where the vision tower, multimodal projector, and language backbone are all unfrozen.
We train the model for 1 epoch with a batch size of 8. The AdamW optimizer is adopted with an initial learning rate of $1\times10^{-5}$, weight decay = 0.01, and a cosine annealing schedule.
A warm-up ratio of 10\% of total iterations is used to stabilize the optimization.
Gradients are clipped with a maximum norm of 1.0. The maximum input length is 8192 tokens.

During the visual search stage at inference, the Monte Carlo Tree Search (MCTS) procedure performs 10 simulations per query with a maximum search depth of 5. The reward function balances semantic verification and localization accuracy via a weighting coefficient of 0.1.

\subsection{Quantitative Results}
\begin{table}[t]
\centering
\footnotesize
\setlength{\tabcolsep}{3pt}
\begin{tabular}{l|cc|cc}
\toprule
\multirow{2}{*}{\textbf{Methods}} & 
\multicolumn{2}{c|}{\textbf{RSVG-HR}} & 
\multicolumn{2}{c}{\textbf{OPT-RSVG}} \\
\cmidrule(lr){2-3} \cmidrule(lr){4-5}
& Pr@0.5  & meanIoU &
Pr@0.5  & meanIoU \\
\midrule
\multicolumn{5}{l}{\textbf{Specialist Models}} \\
\midrule
\color{gray}
MSVG\cite{ding2025visual} & \textcolor{gray}{83.6}  & \textcolor{gray}{72.9} & --  & -- \\
\color{gray}
LQVG\cite{2024Lanrsvghr} & \textcolor{gray}{87.4} & \textcolor{gray}{71.6} & --  & -- \\
\color{gray}
LPVA\cite{li2024optrsvg} & --  & -- & \textcolor{gray}{78.0} & \textcolor{gray}{66.2} \\

\midrule
\multicolumn{5}{l}{\textbf{General VLMs / General MLLMs}} \\
\midrule
GLM-4.1V-9B-Thinking\cite{hong2025glm} & 30.7 & 29.8 & 43.4 & \underline{43.1} \\
Qwen2.5-VL-3B\cite{Qwen2.5-VL} & 29.4  & 32.1 & 24.4  & 25.5 \\				
Qwen3-VL-4B\cite{qwen3vl2025} & \underline{47.7}  & \underline{42.4} & \underline{43.9}  & 42.4 \\
\midrule
\textbf{GeoViS(Our)} & \textbf{51.5}  & \textbf{46.5} & \textbf{70.3}  & \textbf{61.5} \\
\bottomrule
\end{tabular}
\caption{Results on RSVG-HR and OPT-RSVG. GeoViS is trained on Qwen2.5-VL-3B. Best and \underline{second-best} scores are highlighted. Missing values denote metrics unreported by the original papers.}
\label{tab:hr_opt_rsvg_results}
\end{table}

\begin{figure*}[t!]
  \centering
  \includegraphics[width=\linewidth]{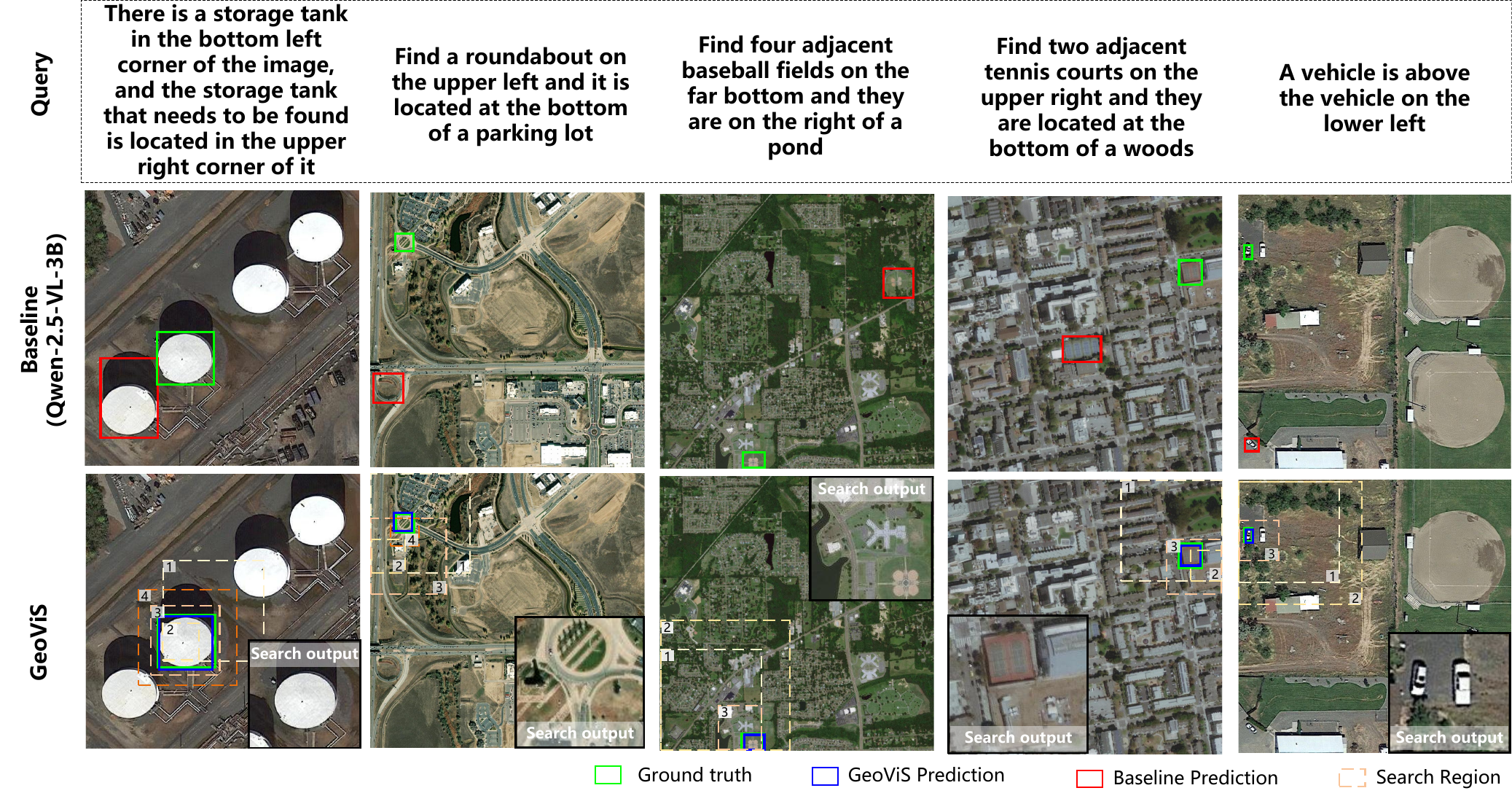}
  \caption{Qualitative results on DIOR-RSVG, OPT-RSVG, and RSVG-HR comparing baseline (Qwen-2.5-VL-3B) with GeoViS. }
  \label{fig:qualitative_results}
\end{figure*}

Tables \ref{tab:dior_vrs_geo_results} and \ref{tab:hr_opt_rsvg_results} report results on five remote sensing grounding benchmarks.
Across all datasets, GeoViS achieves the best overall performance, outperforming general-purpose MLLMs and remote-sensing–specific (RS-specific) MLLMs by a clear margin, and approaching specialist grounding models.

On DIOR-RSVG~\cite{zhan2023rsvg}, GeoViS reaches 79.8\% Pr@0.5, significantly higher than RS MLLMs and nearly +30\% above strong general MLLMs.
On VRSBench~\cite{li2024vrsbench}, GeoViS attains 68.5\% Pr@0.5, surpassing the strongest RS MLLM by over +5\%.
For GeoChat~\cite{kuckreja2024geochat}, GeoViS provides competitive performance (23.7\%), reflecting stable behavior under conversational queries.
On RSVG-HR~\cite{2024Lanrsvghr} and OPT-RSVG~\cite{li2024optrsvg}, GeoViS further achieves 51.5\% and 70.3\% Pr@0.5, respectively, clearly outperforming general MLLMs on both benchmarks.

Overall, the consistent improvements across diverse datasets, modalities, and resolutions confirm that the proposed geospatially rewarded visual search mechanism provides strong and transferable grounding capability in remote sensing imagery.

\subsection{Qualitative Results}
Fig.~\ref{fig:qualitative_results} presents qualitative comparisons across DIOR-RSVG, OPT-RSVG, and RSVG-HR, demonstrating the robustness of GeoViS under diverse scene scales, text complexities, and object sizes. The baseline model (Qwen-2.5-VL-3B), which performs single-step reasoning, often fails to interpret long or compositionally complex queries and consequently mislocalizes the target object—especially when the description relies on geospatial relations or when the target is small relative to the global scene.
In contrast, GeoViS conducts a multi-step visual search that incrementally refines spatial hypotheses. By parsing the textual description into geospatial semantics and integrating the resulting visual cues, GeoViS can accurately identify the correct region before performing final grounding. This progressive search procedure enables the model to handle queries of varying lengths, understand relational and contextual cues, and localize objects across different resolutions and scales. Overall, the qualitative examples illustrate that GeoViS achieves consistently precise grounding where single-step baseline methods struggle.

\subsection{Ablation Study}
\begin{table}[t]
\centering
\begin{tabular}{lcc}
\toprule
Method &  Pr@0.5  & meanIoU \\
\midrule
Qwen-2.5-VL-3B (Global only)  & 71.2 & 62.3\\
Qwen-2.5-VL-3B (Global+Local)  & \textbf{82.9} & \textbf{71.1}\\
\bottomrule
\end{tabular}
\caption{Ablation on effective resolution using DIOR-RSVG. 
``Global only'' trains on full-scene images, while ``Global+Local'' additionally provides cropped regions near the target. The clear gains demonstrate the value of localized visual cues.}
\label{tab:res_ablation}
\end{table}
\begin{figure}[t!]
  \centering
  \includegraphics[width=\linewidth,keepaspectratio]{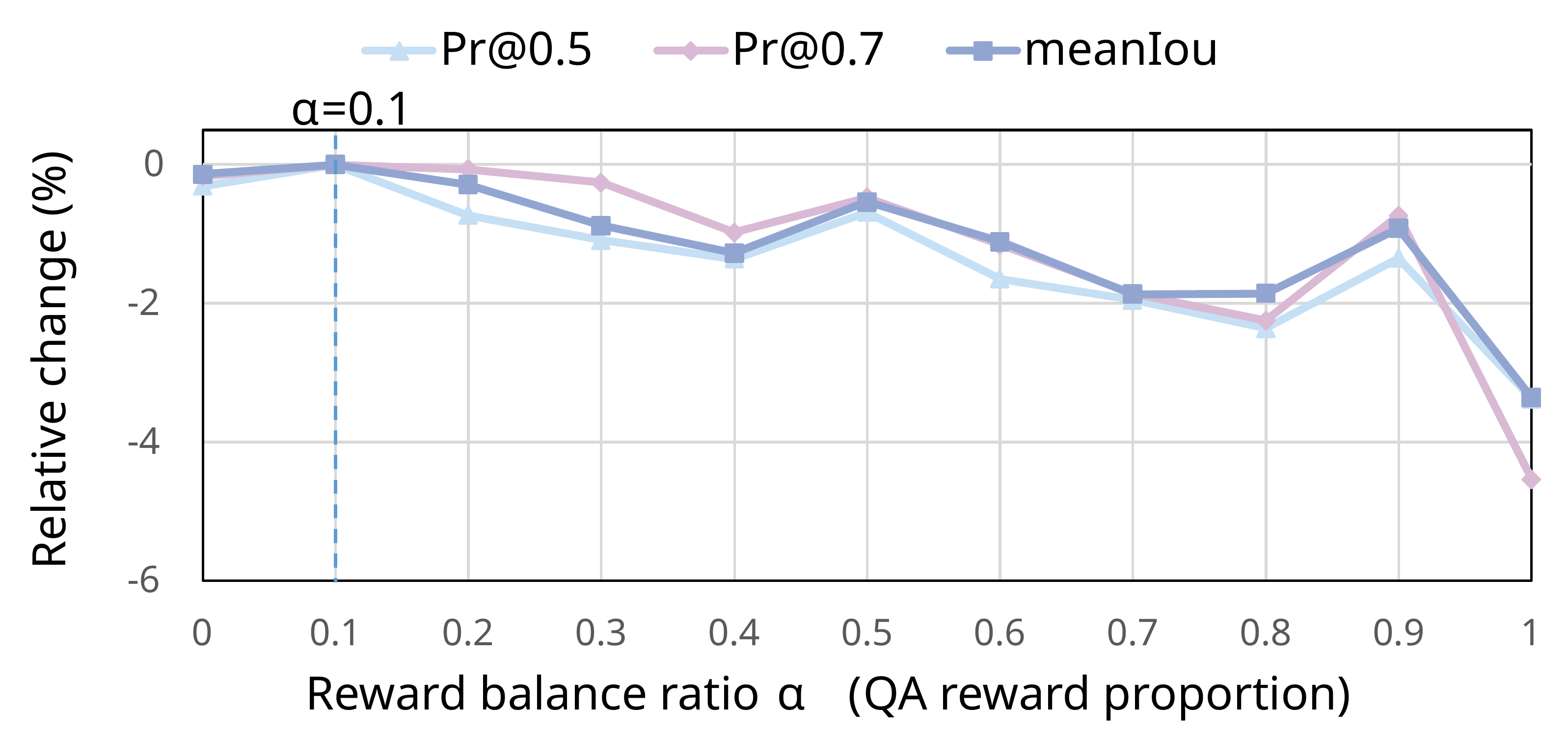}
  \caption{Ablation on the reward balance ratio $\alpha$ on DIOR-RSVG. 
The vertical axis shows the relative performance change (\%) normalized to the maximum value for better visualization.}
  \label{fig:reward_ratio_curve}
\end{figure}

\textbf{Effective Resolution.}
\label{sec:preExp}
Preliminary experiments demonstrate the significant impact of effective resolution on grounding performance.
To verify this, we compare two variants of Qwen2.5-VL-3B trained on the DIOR-RSVG dataset under the same training setup:
(1) using only full-scene images as input, and (2) using both the global image and a local region cropped around the target.
As shown in Table~\ref{tab:res_ablation}, the second configuration achieves notably higher accuracy, indicating that localized visual cues effectively enhance the perceptual resolution of small targets.
This observation motivates our GeoViS framework, which searches for potential local regions containing the target, thereby dynamically increasing effective resolution during grounding.

\textbf{Hyperparameters.} 
We analyze the effect of the reward balance ratio $\alpha$, which controls the proportion of the QA reward $r_{QA}$ in the total reward.
As shown in Fig.~\ref{fig:reward_ratio_curve}, we fix all other settings and vary $\alpha$ from 0 (only $r_{IoU}$) to 1.0 (only $r_{QA}$).
Both reward terms contribute positively to localization accuracy, while excessive emphasis on either term degrades performance.
The optimal result occurs at $\alpha=0.1$, where semantic verification and spatial consistency are properly balanced, confirming that the joint reward is essential for guiding effective visual search.

\textbf{Atomic Operations.}
We further conduct an ablation on the atomic operations in the visual search process, including the \emph{Zoom-in} action and the acquisition of QA and IoU rewards.
As shown in Table~\ref{tab:atomic_ablation}, this experiment is performed on the DIOR-RSVG dataset with the ViT backbone frozen for quick validation.
The baseline uses zero-shot Qwen2.5-VL-3B for all three operations, while subsequent rows progressively replace each step with the trained modules.
Performance improves steadily as more atomic operations are incorporated, demonstrating the effectiveness of decomposing the search process into atomic operations and constructing task-specific training data.
This result verifies that explicitly modeling the search as a sequence of trainable atomic steps provides a more structured and interpretable grounding process.

\textbf{Generalization.} 
We further evaluate the cross-dataset generalization of our model trained using the DIOR-RSVG dataset.
As shown in Table~\ref{tab:gen_ablation}, we compare three settings: (1) zero-shot Qwen2.5-VL-3B, (2) Qwen2.5-VL-3B fine-tuned on DIOR-RSVG, and (3) our GeoViS trained on DIOR-RSVG with data expanded through atomic operations.
When transferred to VRSBench and OPT-RSVG, our method achieves substantially higher precision than both zero-shot and fine-tuned baselines.
This indicates that the learned atomic operations capture transferable visual reasoning patterns, allowing GeoViS to generalize effectively across diverse remote sensing domains.

\begin{table}[t]
\centering
\setlength{\tabcolsep}{5pt}
\begin{tabular}{lcccc}
\toprule
\textbf{Setting} &\textbf{zoom-in}& \textbf{$r_{\mathrm{QA}}$} & \textbf{$r_{\mathrm{IoU}}$} &\textbf{Pr@0.5} \\
\midrule
Baseline & $\times$ & $\times$  & $\times$  & 69.5 \\
+zoom-in & \checkmark & $\times$  & $\times$  & 70.2 \\
+zoom-in + $r_{\mathrm{QA}}$ & \checkmark & \checkmark& $\times$  & 71.2  \\
+zoom-in + $r_{\mathrm{IoU}}$ & \checkmark & $\times$ & \checkmark & 73.2\\
Full model & \checkmark & \checkmark & \checkmark & \textbf{74.5} \\
 \bottomrule
\end{tabular}
\caption{Ablation of atomic operations on the DIOR-RSVG dataset. Each component provides a measurable gain.}
\label{tab:atomic_ablation}
\end{table}

\begin{table}[t]
\centering
\small
\renewcommand{\arraystretch}{1.1}
\begin{tabular}{l|cc|cc}
\toprule
\multirow{2}{*}{\textbf{Methods}} & 
\multicolumn{2}{c|}{\textbf{VRSBench}} & 
\multicolumn{2}{c}{\textbf{OPT-RSVG}} \\
\cmidrule(lr){2-3} \cmidrule(lr){4-5}
& \textbf{Pr@0.5}  & \textbf{meanIoU} &
\textbf{Pr@0.5}  & \textbf{meanIoU} \\
\midrule
Zero-shot  &37.3&35.3& 29.4&32.1  \\
Fine-tuned & 39.3&37.7 & 33.6&32.7  \\
Ours & \textbf{49.0}&\textbf{44.7} & \textbf{41.0}&\textbf{40.6}  \\
\bottomrule
\end{tabular}
\caption{Cross-dataset generalization results. 
Models are trained on DIOR-RSVG and evaluated on VRSBench and OPT-RSVG. GeoViS demonstrates strong transferability across datasets, outperforming fine-tuned baselines by a clear margin.}
\label{tab:gen_ablation}
\end{table}

\section{Conclusion}
In this work, we introduced GeoViS, an Active Geospatial Search framework that reformulates remote sensing visual grounding as a two-stage process of hierarchical search and conditional grounding.
By integrating a unified Visual Reward-Action-Grounding (VisualRAG) model that provides multimodal guidance for exploration and localization, GeoViS achieves interpretable, step-wise reasoning across large-scale geospatial imagery.
Extensive experiments on multiple benchmarks demonstrate its strong performance and cross-dataset generalization, highlighting GeoViS as a general and transferable paradigm for visual grounding in remote sensing.

{
    \small
    \bibliographystyle{ieeenat_fullname}
    \bibliography{main}
}

\end{document}